\documentclass{layout/siccs-thesis}
\usepackage{parskip} %Automatic blank lines if there is an enter
\usepackage{amsmath}
\usepackage{booktabs}
\usepackage{epigraph}
\usepackage{setspace}
\usepackage{multirow}
\usepackage{array}  
\usepackage{float}  
\PassOptionsToPackage{table}{xcolor}  % 或你需要的其他选项，比如 {table,dvipsnames}
\usepackage{xcolor}
\usepackage{amssymb} 
\usepackage{subcaption}
\usepackage{siunitx}   
\usepackage{bbding}
\usepackage{pifont}
\usepackage{svg} 
\usepackage{adjustbox} 
\usepackage[style=numeric, backend=biber]{biblatex}
\DeclareDelimFormat{nameyeardelim}{\addcomma\space}

\DeclareSIUnit\equivalents{eq}
\DeclareSIUnit\water{kgw}
\DeclareSIUnit\year{yr}
\DeclareSIUnit\dollar{\$}

\setlist{itemsep=-2pt} % Reducing white space in lists slightly
\begin{document}

\frontmatter

%% Defining the main parameters

\title{Mitigating Spurious Correlations in Weakly Supervised Semantic Segmentation via Cross-architecture Consistency Regularization}
% \subtitle{Industrial Exhaust Smoke emission-Oriented Pseudo label Refinement Method}

\author{Zheyuan Zhang}
% \studentnumber{(15321495,2797576)}
% \mastertrack{FCC}
% \affiliation{University of Amsterdam
%             Vrije Universiteit Amsterdam}
% \companysupervisor{Your company supervisor}
% \firstsupervisor{Prof. Yen Chia Hsu}
% \secondsupervisor{Prof. Nanne van Noord}
% \handindate{\monthname[\the\month] \the\day, 2025} % Automatic date formatting

\definecolor{title}{HTML}{4884d6} 
% Color for title

% title page
% ====== frontmatter/_titlepage.tex ======
\title{Mitigating Spurious Correlations in Weakly Supervised Semantic Segmentation via Cross-architecture Consistency Regularization}

\author{
    Zheyuan Zhang \\
    University of Amsterdam  \\
    mickey.zhang@student.uva.nl
    \and
    Yen-Chia Hsu \\
    University of Amsterdam \\
    y.c.hsu@uva.nl
    % \and
    % Second Reader Name \\
    % Vrije Universiteit Amsterdam \\
    % second@example.com
}

\date{}
\maketitle

% \newpage
% \thispagestyle{empty}
% \vspace*{1cm} % 距离顶部的空白

% \begin{center}
% % 横线
% \rule{0.6\textwidth}{0.4pt} % 第一个参数是宽度，第二个是线条粗细

% \vspace{1cm} % 横线与句子之间的间距

% \begin{minipage}{0.8\textwidth}
% \centering
% \itshape
% ‘Going deeper. And still deeper. The green mountains.’\\[1em]
% \textit{-Santoka Taneda}
% \end{minipage}
% \end{center}

% \newpage

% abstract
\chapter*{Abstract}
%创建一个名为 “Abstract” 的章节标题，* 表示这个章节不会被编号。因此，在文档中不会出现 “Chapter 1: Abstract” 这样的标题，而是直接显示 “Abstract”。
\addcontentsline{toc}{chapter}{Abstract}

% Abstracts are usually around 100–300 words
\noindent 
The precise monitoring and localization of industrial exhaust smoke emissions is critical for effective environmental regulation and public health protection.
Mainstream methods rely on fully supervised setting to build the segmentation model, demanding a large amount of data with labor-intensive and costly pixel-level annotations.~Nevertheless, scarcity of pixel-level labels is a significant challenge in practical scenarios.~In specific domains like industrial smoke or medical imaging, acquiring such detailed annotations is particularly challenging, often requiring specialized knowledge for accurate annotation.~To alleviate the issue of lacking pixel-level annotations, one promising approach is weakly supervised semantic segmentation~(WSSS), which uses weaker supervision information like image-level to minimize the need for fine-grained annotations.

However, because of the supervision gap and  inherent knowledge bias, existing WSSS methods that rely solely on class activation maps (CAMs) generated from image-level labels suffer from  several crucial limitations: incomplete foreground coverage, inaccurate object boundaries, and spurious correlations arising from co-occurrence context information, resulting in low-quality pseudo-labels.~In our domain specific task, the co-occurrence problem is particularly pronounced, as industrial emission are always spatially coupled with chimneys.

Previous approaches mainly rely on additional knowledge or human prior to mitigate spurious correlations,~suffering from insufficient  extraction of comprehensive semantic information.~In addition,~such strategies are difficult to scale and overlook the root cause of the problem--the model’s inherent bias toward co-occurring contextual cues.~In this work, we observe that there is an inconsistency between CAMs generated with different network architectures, which can be leveraged to suppress irrelevant context and reduce knowledge bias.

To this end, we propose a novel and efficient WSSS framework that directly tackles the co-occurrence issue without relying on external supervision.~In contrast to previous approaches, which typically rely on a single architecture network, we adapt the teacher-student framework to combine the fine-grained spatial modeling capability of ViTs with the precise localization strength of CNNs to solve the co-occurrence issue,~offering complementary semantic guidance.~To achieve this, we introduce a knowledge transfer loss function that promotes cross-architecture consistency by aligning the internal representations of ViTs and CNNs.
This enables the networks to fully preserves both spatial structures and semantic information, encouraging the classifier to focus on foreground regions more accurately.~Further, we also  incorporate advanced post-processing techniques to address partial coverage problem to further improve the quality of pseudo-masks.
Extensive experiments on a custom and domain-specific dataset reveal the effectiveness of our proposed framework and the superiority over previous methods.~The code is available at \url{https://github.com/zhang-mickey/Master-thesis}.

% Given the aforementioned limitations,this work focuses on improving the feature representations learned by classifiers to constrain the CAM to better separate foreground and background cues. 
% \lipsum[2-4]

\tableofcontents
%\listoffigures
%\listoftables

\chapter*{Nomenclature}
\addcontentsline{toc}{chapter}{Nomenclature}

\section*{Abbreviations}

\begin{longtable}{p{2.5cm}p{8cm}}
    \toprule
    Abbreviation & Definition \\
    \midrule\endhead % Abbreviations added alphabetically here:
    WSSS & Weakly Supervised Semantic Segmentation\\
    SAM & Segment Anything Model \\
    KD & Knowledge Distillation \\
    ViT&Vision Transformer\\
    CAM&Class Activation Map\\
    CRF&Conditional Random Field\\
    CNN&Convolutional Neural Network\\
    CLIP&Contrastive Language-Image Pre-training\\
      \bottomrule
\end{longtable}
% \section*{Variable Names}

% \begin{longtable}{p{2.5cm}p{8cm}
%     \toprule
%     Symbol & Definition \\
%     \midrule\endhead % Latin symbols added alphabetically here:
%     % $\alpha$ & greek alpha & [\si{\newton}] \\
%     % \midrule
%     % $\beta$  & greek beta  & [\si{\joule}]   \\
%     % \bottomrule
%     $\lambda$  & weighting factor of transfer loss   \\
%     \bottomrule
% \end{longtable}

%% Arabic page numbering
\mainmatter

\chapter{Introduction}
Unlike objects with well-defined contours,~smoke exhibits inherent characteristics~(transparent appearance, low contrast) that pose unique challenges for segmentation.
While fully supervised learning methods have achieved impressive results,~these methods require a large scale of training images with pixel-level annotations, which is expensive and time-consuming, making it hard to be used in real world scenarios.~Therefore, to avoid pixel-level labeling, weakly supervised semantic segmentation has emerges as a effective solution, which typically rely on various forms of weaker supervision, such as image-level labels~\cite{xu2022multi,peng2023usage,rong2023boundary,tang2024hunting,lee2022weakly,choe2019attention,zhang2023weakly,wei2017object,zhou2022regional,li2023transcam,zhu2023weaktr,chen2023extracting}, bounding box~\cite{dai2015boxsup,papandreou2015weakly,hsu2019weakly,oh2021background}, scribbles~\cite{zhang2021dynamic,wang2022cycle} and points~\cite{fan2023toward}.~Notably, our work adopts the mainstream approach of using only image-level labels for supervision,~as image-level labels can be collected quite efficiently in real-world scenarios and it's the most challenging one.~For image-level annotations, class activation map is usually utilized  to generate pseudo masks.~CAM~\cite{zhou2016learning}  captures the most important pixels of the input image that make a difference in determining classification.~The highly-activated regions, also known as seeds,~provide localization for foreground and are used to generate pseudo masks.~However, there is still a significant performance gap between WSSS with image-level label and fully supervised semantic segmentation.~In this extreme setting, a naive weakly-supervised segmentation model will typically yield poor localization capacity.~This disparity arises because image-level labels do not explicitly provide any spatial information such as the location, shape, and size of the foreground in the image.
And the features learned in classifier are mainly used for the classification task rather than dense predictions.

\begin{figure}[H]
    \centering
    \includegraphics[width=1\linewidth]{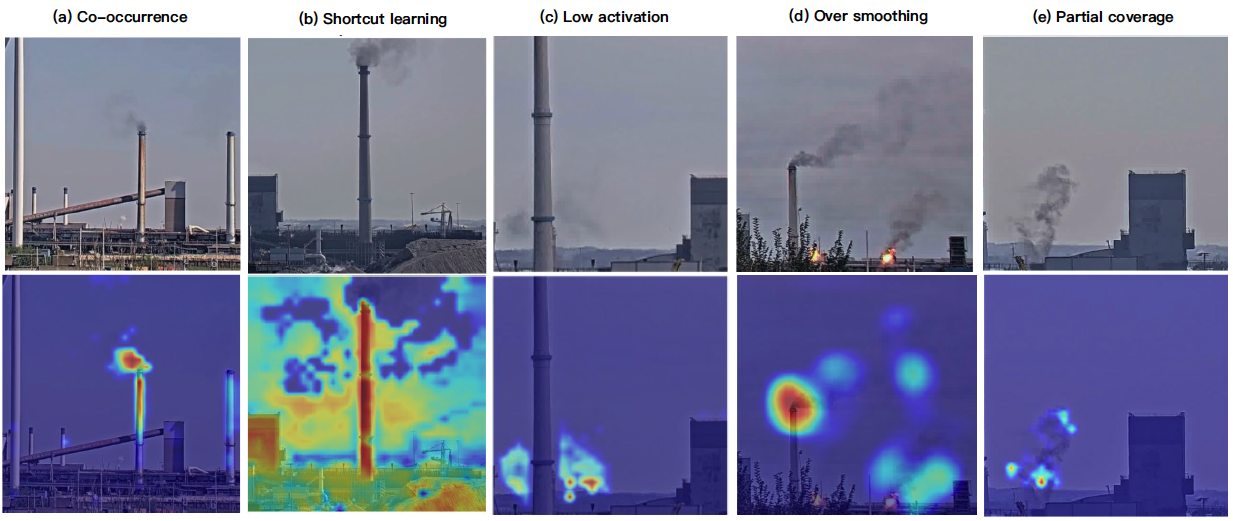}
    \caption{CAMs supervised with image-level labels often suffer from issues like co-occurrence  and partial activation.}
    \label{fig:introduction}
\end{figure}

By our experiment, one observation is that the classifier achieves high accuracy but doesn't focus on the semantically meaningful parts of the image (the foreground smoke), instead focusing on background features or environmental cues.~As shown in Figure~\ref{fig:introduction}, by our experiment,~CAMs suffer from issues like co-occurrence and partial coverage.~For smoke segmentation, in the context of weakly supervised learning, it is harder to distinguish smoke from the complex background due to the intrinsic similarity,~resulting in inaccuracy boundaries.~CAMs can also wrongly activate large fuzzy areas and confuse background haze or gray cloud with foreground smoke,~which leads to high false positives,~especially in  blurry backgrounds.~Meanwhile, there exists a common spurious correlation problem:~chimneys and smoke, making it hard for the classifier to determine the location of smoke.~Furthermore, optimal CAM threshold varies significantly across images, fixed threshold  value for converting CAMs to pseudo masks produces low-quality pseudo masks.~Lastly,~classifiers trained on biased datasets tend to do shortcut learning~\cite{geirhos2020shortcut}, which means networks exploit easy-to-learn global image cues instead of learning smoke-specific features.~From the above analysis, we identify three critical challenges in weakly supervised smoke segmentation with image-level labels for smoke segmentation: co-occurrence, inaccurate
object boundaries and partial activation, which can all be addressed by our framework.

% One observation is that the classification model discovers shortcuts to classify images by memorizing background patterns.

% In this work, we aim to fill these gaps in the contrastive learning literature in the context of segmentation of

% The challenges for weakly smoke semantic segmentation are as follows:
      
% CAMs often capture incomplete foreground regions with poor coverage of subtle smoke regions in CAMs.

% Classifier predictions heavily dictate activation region quality.Mislocalized pseudo-labels degrade model performance.
% Error amplification through CAM-to-mask conversion pipeline

%     \item[3. Shortcut Learning] 
%  which results in false activation of non-smoke objects (e.g., sky and building) and compromised generalization to novel environments
% \end{description}

% \begin{figure}[H]
%     \centering
%     \includegraphics[width=0.7\linewidth]{challenges.png}
%     \caption{CAMs supervised with image-level labels often suffer from issues like co-occurrence problem and partial activation.}
%     \label{fig:introduction}
% \end{figure}

Basically, there are mainly two directions to alleviate these shortcomings: one is to improve the classifier to enhance representation learning~\cite{yang2025more,rong2023boundary,tang2024hunting,lai2024weakly,wang2020self,wei2017object,chen2023extracting,he2023mitigating}.~The other one is to use post-processing techniques~\cite{chen2023segment,wu2024wps,ahn2018learningpixellevelsemanticaffinity} to refine the cam and pseudo label.~Post-processing techniques are proposed to address inaccurate boundary and partial coverage problems,~which mainly depend on the RGB and position priors of pixels.~While post-processing can refine CAM and pseudo label to a large extend, it do not have the capacity to deal with co-occurrence and wrongly activated region.~When co-occurring pattern activated by CAMs, applying post-processing techniques will amplify these errors and degrade the quality of pseudo masks.

To address co-occurring issue, previous methods has been proposed like data augmentation~\cite{su2021contextdecouplingaugmentationweakly,xu2022boat}, additional knowledge or supervision~\cite{xie2022clims,lee2021railroad}, prototype learning~\cite{tang2024hunting}, causal inference~\cite{chen2022c} and human prior~\cite{lee2022weakly,hsu2019weakly} to decouple the foreground from the context information.~These approaches aim to decouple the foreground from its co-occurring background or to establish causal relationships between semantic concepts and visual cues.
However, these approaches often rely on additional supervision, domain-specific knowledge.~Although~\cite{tang2024hunting} uses consistency learning to alleviate the knowledge bias, it is based on single architecture network and introduces extra supervision signals.~In our work,~we propose a different way to mitigate spurious correlations without extra supervision or prior knowledge.

% Despite the remarkable strides achieved by these studies, a significant challenge remains.

% As there is no pixel-level supervision in weak supervision, CAM is only responsible for classification,there is no doubt that this would occur.
% this locality is due to the fact that
% CAM is extracted from a discriminative model. The training of such model naturally discards the non-discriminative
% regions which confuse the model between similar as well as
% highly co-occurring object classes\cite{chen2023extracting}.

\begin{figure}[H]
    \centering
    \includegraphics[width=1\linewidth]{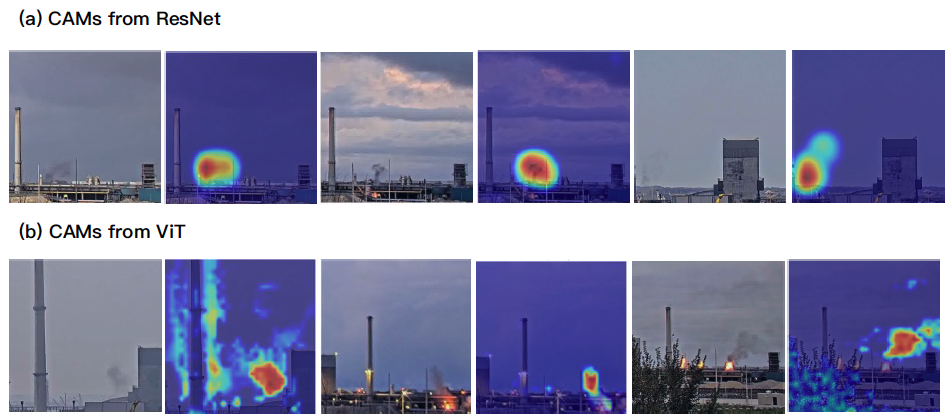}
    \caption{CAMs generated from different backbone.}
    \label{fig:resnet_vit}
\end{figure}

% Our work is also loosely related to \cite{},which transfer knowledge between global view network with local view network.
Another drawback of the aforementioned methods is that they  focus only on a single architecture model(e.g., CNN or ViT), which inherently limits their ability to capture both local spatial details and global semantic context.~As mentioned by~\cite{raghu2021vision}, there exists clear differences in the internal visual representation of ViT and CNN, which inspired us to generate CAMs using networks with distinct architectures.
By our experiment, one key observation is that ViT and ResNet generate distinct CAMs, which is also highlighted by~\cite{peng2023usage}. ~As shown in Figure~\ref{fig:resnet_vit}, the CAMs generated by ResNet shows good localization but suffering over smoothing at boundaries issue.~While ViT captures shaper CAMs with clear edge information but suffering from spurious corelation issue.~As demonstrated in~\cite{sun2024massive},~due to self-attention mechanism designed for modeling long-range dependencies,~ViT tends to aggregate global semantics even in patches with low information.~This kind of behavior is beneficial for capturing holistic object representations but may also results in  spurious activations in weakly supervised settings.~Therefore,~ as shown in Figure~\ref{fig:resnet_vit},~ViT tends to suffer from incorrect activations.~In contrast, CNN is better at distinguishing  the foreground from its co-occurrence object, but often generate over-smoothed CAMs that lack precise boundaries.~Based on this core observation, a natural idea arises: can the knowledge from ResNet~(ViT) complement the deficiencies of ViT~(ResNet) without compromising their strengths?
% ViT is able to generate CAMs with clearer object boundaries, which CNNs often struggle with.
% Most CNN-based methods have the problem of over-smoothing at boundaries

Motivated by above analysis and observation, we propose a teacher-student framework to tackle the co-occurring issue by enforcing cross-architecture consistency between two heterogeneous networks.~In addition,~to facilitate mutual understanding between the networks, we introduce a  knowledge transfer  loss function for feature alignment, which enables more effective knowledge sharing.~Unlike mainstream approaches that rely on additional supervision or human prior, our work addresses the co-occurring issue from a new perspective without introducing extra knowledge.~Instead, our method leverages the rich and complementary semantic representations inherently learned by diverse network architectures.~Based on the above observation and analysis, we want to answer the following questions:
\begin{enumerate}[label=\textbf{RQ\arabic*}, wide=0pt, leftmargin=2em]
    \item Why does the classifier achieve very high accuracy but the activated region of CAM is not accurate or even fail to localize the foreground?
    \item Is it possible to address co-occurrence issue without external supervision or additional knowledge?
    \item Can we collaboratively  aggregate heterogeneous features from CNN based and Vision Transformer based models to address co-occurrence issue?
\end{enumerate}

% \begin{figure}[H]
%     \centering
%     \includegraphics[width=0.7\linewidth]{figures/resnet_vit.png}
%     \caption{CAMs generated by different backbone.The first and second line is for ResNet and ViT respectively.}
%     \label{fig:resnet_vit}
% \end{figure}

Collectively,~the main contributions can be summarized as follows:

(1)~Our empirical analysis reveals complementary strengths between CNNs and ViT in CAMs  generation.
Building on this insight, we propose a straightforward yet effective  WSSS framework  based on the teacher-student paradigm to reduce knowledge bias and mitigate spurious correlations.
By enforcing cross-architecture consistency at the feature level, our approach integrates the discriminative localization of CNNs with the boundary knowledge of ViTs.
This facilitates segmentation-aware training within a classification setting, which is not well explored.
To the best of our knowledge, this is the first attempt to effectively leverage the distinct advantages of ViT and CNN through  consistency regularization at feature level to address the co-occurrence problem in WSSS using only image-level labels without additional supervision.

(2)~To further improve the quality of pseudo masks,we implement and systematically evaluate a bunch of advanced post-processing methods.
Our findings reveal that several methods significantly outperform others in the context of smoke segmentation under WSSS, providing  insights for refining pseudo-label in domain-specific scenarios.

(3)~We evaluate our framework on a custom dataset, validating the efficiency of our method and the superiority over previous methods.

% In contrastive learning, selecting anchor, positive, and negative samples properly is crucial for effective representation learning.
% The work shows that sampling is important exclude 
%  a suitable loss function. \cite{wu2018samplingmattersdeepembedding}.Many different loss functions perform similarly under a good sampling strategy.
 
% A positive pair often consists of image augmentations of the sample, and negative pairs are formed by the anchor and randomly chosen samples.
% In practice,constrastive learning methods benefit from a large number of negative samples.

% A rich line of work focuses  the role of positive pairs.This work propose the hard negative samples.
% \cite{robinson2021contrastivelearninghardnegative}

% Given an anchor $a$ and a positive example $p$,obtain a negative instance $n$ via 

% Batch construction also matters.In order to obtain more information,a larger batch size is prefer.

% the discriminative areas highlighted by CAMs are in turn used as seeds that will be propagated to cover the entire object area.
% \cite{zhou2016learning}.

% For well-calibrated  bespoke model.

\chapter{Related Work}
In this section, we will first review the existing WSSS methods using image-level labels, then move a step further to review the relevant works on how to refine the pseudo masks.

\section{Weakly Supervised Semantic Segmentation}

\begin{figure}[H]
    \centering
    \includegraphics[width=0.95\linewidth]{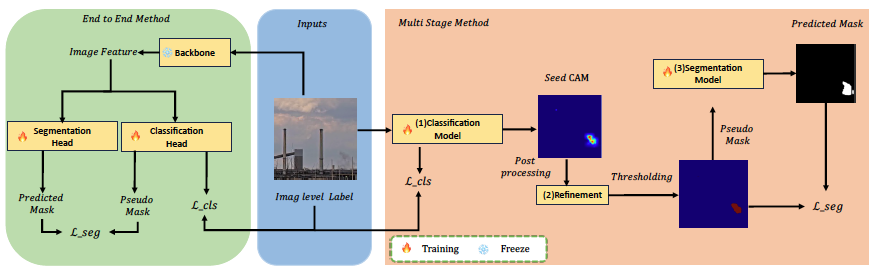}
    \caption{The pipeline of multi stage and end to end method in WSSS.~In our framework,we adopt multi stage pipeline.}
    \label{fig:stage}
\end{figure}
% \begin{figure}[H]
%     \centering
%     \includegraphics[width=0.95\linewidth]{architecture.png}
%     \caption{The pipeline of multi stage and end to end method in WSSS.In this work,we only adopted multi stage paradigm.}
%     \label{fig:stage}
% \end{figure}

\subsection{WSSS In General}
Various WSSS methods are proposed to get rid of expensive pixel-level annotations, which can be categorized as multi-stage method~\cite{rong2023boundary,wei2017object,ahn2018learningpixellevelsemanticaffinity} and end to end method~\cite{he2024progressive,xu2023self,ru2022learning,araslanov2020single,wu2024dino,yang2025more}.~As shown in Figure~\ref{fig:stage}, the standard pipeline of multi-stage WSSS is (1) training a classifier with image-level annotations for obtaining CAMs(seed generation), (2) refining the CAMs into pseudo-labels(mask generation), and (3) training a semantic segmentation model using the pseudo-labels.~Phase (2) is where post-processing techniques like AffinityNet~\cite{ahn2018learningpixellevelsemanticaffinity} can be applied.~During pseudo-mask generation stage, a fixed threshold is used to separate the foreground from background.~End-to-end WSSS pipeline aims to streamline multi-stage into a single training phase, which is more efficient but challenging.
Most existing methods rely on Class Activation Maps to derive pixel-level pseudo-labels and use them to train a fully supervised semantic segmentation model.~There are also multiple-instance learning based method, which is beyond our discussion scope.~Meanwhile, due to the inherent limitations of vanilla CAM,~numerous variants have been proposed to improve its localization quality and feature representation like Grad-CAM~\cite{selvaraju2017grad},~Score-CAM~\cite{wang2020score},~LPCAM~\cite{chen2023extracting} and ReCAM~\cite{chen2022class}.

\subsection{WSSS with only image-level labels}
In practical scenarios, image-level annotations are the most accessible form of supervision,~making them a popular choice for WSSS.~However, the initial CAMs generated with image-level labels are coarse and noisy, frequently suffering from partial object coverage, inaccurate boundaries, and contextual bias caused by spurious correlations.~To address these challenges, numerous studies have been conducted.Depending on the specific issues they aim to solve, these methods can be broadly categorized into three different groups:
% However,there is still a gap between models trained with image-level labels and models trained with pixel-level masks.
\textbf{(1)Partial coverage:} Partial coverage is a long-standing challenge for weakly supervised semantic segmentation.~Earlier works in weakly supervised semantic segmentation mainly focus on solving partial activation issue and expand object coverage to capture the full extent of objects by iterative erasing~\cite{wei2017object}, stochastically ensembling score maps~\cite{lee2019ficklenet} and affinity based method~\cite{ahn2018learningpixellevelsemanticaffinity}.
While these methods show their successes, they fail to determine accurate object boundaries of the target object because they have no semantic clue to guide the object’s shape.~Most recently, using prior knowledge from VLM model and vision foundation model assisted weakly supervised semantic segmentation  has emerged as a trend~\cite{zhang2024frozen,xie2022clims,kweon2024sam,chen2023segment}.~These methods are based on the hypothesize that the quality of  pseudo masks can be refined by appropriately integrating initial seed from CAM and the the abundant knowledge from use off-the-shelf model like vision language model~(CLIP) and vision foundation model~(SAM).

\textbf{(2)Inaccurate boundaries:} Due to the intrinsic property that the CAM is responsible for classification, it is non-trivial to balance the recall of the foreground and the false-positives of the background.
Thus, there are some approaches focus on improving the object boundaries of pseudo masks~\cite{fan2020learning,chen2020weakly,rong2023boundary}.~For example, \cite{rong2023boundary}proposes a co-training method for training the segmentation model.

\textbf{(3)Co-occurrence issue:} In addition,CAM often fails to distinguish the foreground from its co-occurred background,which is hard to be separated without additional information.~Existing works mainly address this  by introducing external supervision~\cite{lee2021railroad,xie2022clims} ,~causal inference~\cite{chen2022c} and human priors~\cite{lee2022weakly,hsu2019weakly}.
For example,~\cite{lee2021railroad}devise a joint training strategy to fully utilize the complementary information from image-level  labels and saliency map.
Recently, to address co-occurrence issue,~some works~\cite{yin2024fine,yang2024separate} leverages contrastive learning to enhance semantic representation.
However,~this kind of method is computational expensive, demanding memory bank to store positive and negative pairs for better performance.~To this end, we observe that rich semantics is inherently encoded within heterogeneous architectures, which helps differentiate co-contexts. Therefore, we design a cross-architecture knowledge transfer loss via consistency regularization to foster the classifier to learn complementary features.
To the best of our knowledge, we are the first to report that leveraging cross-architecture consistency can effectively suppress co-occurrence bias during the representation without the need of external supervision.

% Existing research also points out that the bottleneck of WSSS comes from the global threshold\cite{lee2022threshold},and thus build a robust model against threshold choice.
% ~\cite{du2022weakly} come up with a pixel-to-prototype contrastive learning method to narrow the gap between classification and segmentation.
% 
% WSSS with \textbf{Additional Source of Information}:
% There are also approaches that combine other sources of knowledge to enhance the guidance like Saliency map~\cite{lee2021railroad}.

% WSSS with \textbf{contrastive learning}:
% Contrastive learning has emerged as a powerful tool for weakly supervised semantic segmentation, enabling models to distinguish foreground from background through inter-image relationships.Recent works~\cite{zhou2022regional} incorporate cross-image contrastive learning to enhance pseudo-label quality.

% Grad-CAM utilizes the gradients of the target class flowing into the final convolutional layer to weight the importance of feature maps, offering a class-specific localization map that is more interpretable. Score-CAM, in contrast, eliminates the reliance on gradients by using the model’s output scores as weights, thus improving robustness and avoiding noisy gradient signals.
% These variants differ primarily in how they compute the importance of each feature map or spatial location

% Several post-process methods based on deep learning have also been proposed.

% This research propose a new seeding loss function
% \cite{kolesnikov2016seedexpandconstrainprinciples}.
% In this section,we provide a technical description of  the approach.

\section{Knowledge Distillation}
In order to extract richer and sufficient semantic information, some previous works \cite{xu2023self,} introduce knowledge distillation into WSSS to refine pseudo labels without introducing external
supervision.~However, most existing  methods still rely on a single type of backbone.~While a few works~\cite{park2024precision,li2025pdseg} in WSSS  utilize both ResNet  and ViT to improve the quality of CAMs, their knowledge transfer strategies are simplistic or rely on additional supervision, lacking a deeper exploration of effective cross-architecture knowledge transfer.~For example, \cite{park2024precision} fuses CAMs from ViT and ResNet using simple ensemble operations such as OR and AND, without explicitly modeling inter-architecture feature alignment.~Similarly, \cite{li2025pdseg} performs patch-level distillation rather than pixel-wise transfer, which limits the granularity and effectiveness of the knowledge transfer, suffering from
insufficient extraction of comprehensive feature.~To the best of our knowledge,~no  prior work has effectively integrated CNNs and ViTs within a WSSS framework to explicitly address the co-occurrence issue. ~In our work, we introduce a cross-architecture knowledge transfer loss to facilitate fine-grained feature transfer between heterogeneous models. Additionally, we incorporate consistency regularization to align and enhance cross-architecture features, further improving the quality of the generated pseudo labels.

In the meantime, some existing works in related domains have already leverage both CNN and ViT together using advanced knowledge distillation techniques~\cite{hinton2015distilling} to enhance representation learning.~For example,~\cite{zhu2023good} explored an online KD paradigm and proposed heterogeneous feature distillation by mimicking heterogeneous features between CNNs and ViT.~\cite{zhang2023weakly} proposed a self-dual teaching network architecture.~cite{shu2021channel} proposed a channel-wise knowledge distillation paradigm.
DeiT~\cite{touvron2021training} shows that using convnet as the teacher model is better than using only the ViT models. ~\cite{zheng2024transformer} proposed a semi-supervised learning framework that leverages Transformer-CNN Cohort  for knowledge transfer between the two students.~\cite{d2021convit} digs into designing hybrid ViT-CNN models, which can be regraded as a different way of ViT-CNN knowledge sharing.~\cite{wang2022cnn} combines the feature-learning strength of CNN and ViT and enables consistency-aware supervision in a semi-supervised manner.~\cite{zhao2023cumulative} proposes a spatial knowledge distillation by leveraging CNN’s local inductive bias.~\cite{ngo2024learning} design a hybrid method to fully take advantage of both ViT and CNN.~\cite{kweon2024sam} transfers the knowledge of SAM to the classifier during the training process,which is similar to our approach.~In our work, we also want to transfer the localization knowledge of ResNet to ViT classifier by imposing cross-architecture consistency using the knowledge transfer loss.
% In particular, the Conformer~\cite{peng2021conformer} utilizes a CNN branch and a transformer branch to fuse local features and global representations interdependently at multiple scales.

% The standard knowledge distillation paradigm~\cite{hinton2015distilling} is a model compression method proposed to learn a effective student model under the guidance of a high-capacity teacher model.
% Recent studies focus more on attention or feature distillation~\cite{heo2019comprehensive}.
% In terms of supervision,the common KD processes are usually performed under full supervision, where the knowledge from the teacher model is perfect and accurate. Unfortunately, in WSSS, the teacher network can only be trained under weak supervision.
% Existing online KD methods for classification employ a ‘Dual-Student’ framework (without the pre-trained model) by enabling the students to learn from each other in a one-stage learning manner
% Existing work already attempts to combine the strengthen of both ViT and CNN.

\section{Consistency Learning}
Considering there is a large supervision gap between fully and weakly supervised semantic segmentation, to address inaccurate segmentation boundaries and partial activation issues, some prior works~\cite{fang2023weakly,sun2023all,du2022weakly,xu2022boat} start to use consistency learning to construct internal supervisions to  refine the pseudo labels.
Consistency regularization~\cite{ouali2020semi} has shown great potential in learning discriminative representations without the need for dense annotations.
This approach is based on the hypothesis that features should retain semantic consistency across different views of the same image.or across different models.
Specifically, \cite{sun2023all} uses augmentation invariant consistency to refine CAMs.~\cite{chen2021semi} imposes consistency regularization on two segmentation networks perturbed with different initialization for the same input image.
SEAM ~\cite{wang2020self} proposed
consistency regularization on predicted CAMs from various transformed images for self-supervision learning.
In the meantime, some previous works  have explored minimizing discrepancies between models of the same architecture.~For example, ~\cite{rong2023boundary,jiang2022l2g} using co-training strategy to construct two parallel networks that share the same architecture to impose consistency in the predictions of the two networks for uncertain pixels.
~\cite{wu2024dupl} propose a dual student network via a discrepancy loss with an adaptive noise filtering strategy, also with the same architecture.
~\cite{peng2023usage} build cross-view spatial activation distribution consistency across different architectures by applying dropout adjustment to the classification models,  but their method lacks feature-level alignment and does not explicitly address whether it can handle spurious correlations.~However, these methods focus on partial coverage problem or cross-view consistency, without considering using cross-architecture consistency to extract abundant unbiased semantic knowledge or tackling the co-occurrence bias.
Inspired by this cross-view consistency methodology, we revisit the problem of spurious correlations and propose to mitigate it by enforcing cross-architecture consistency across diverse architectures without relying on extra supervision.

\chapter{Methodology}
In this section, we first  describe the motivation and overview of this work.~Then we elaborate on the details of the proposed method and final training objective.~Next,~post-processing techniques are introduced.

\begin{figure}
    \centering
    \includegraphics[width=1\linewidth]{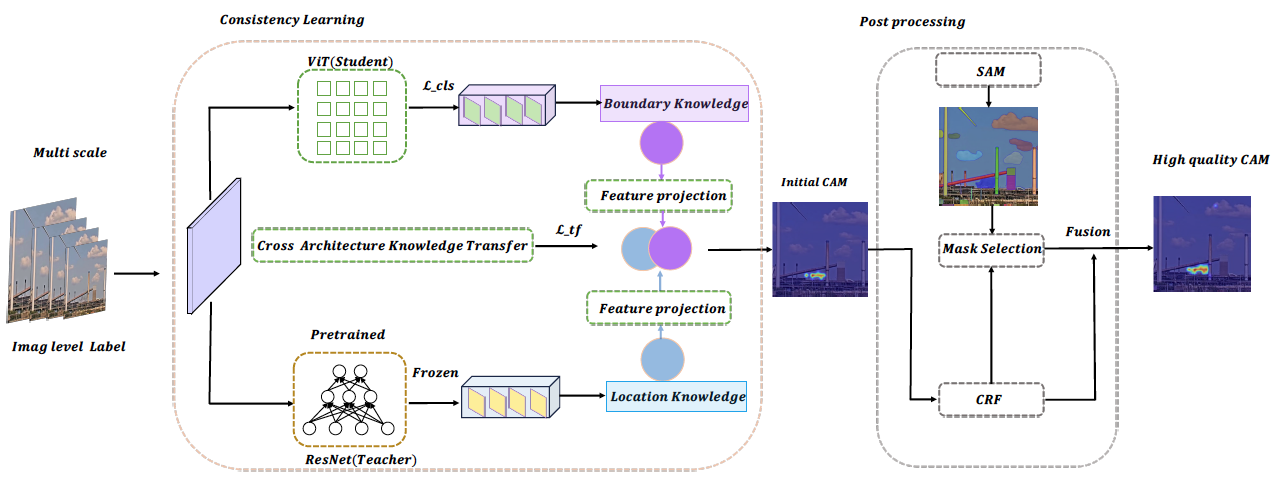}
    \caption{The overall architecture of our proposed framework.}
    \label{fig:overview}
\end{figure}
% \begin{figure}
%     \centering
%     \includegraphics[width=1\linewidth]{figures/overview_fin.png}
%     \caption{The overall architecture of our proposed framework.}
%     \label{fig:overview}
% \end{figure}

% \begin{figure}
%     \centering
%     \includegraphics[width=1\linewidth]{figures/overview_fina.png}
%     \caption{The overall architecture of our proposed framework.}
%     \label{fig:overview}
% \end{figure}
% \begin{figure}
%     \centering
%     \includegraphics[width=1\linewidth]{figures/overview_finall.png}
%     \caption{The overall architecture of our proposed framework.}
%     \label{fig:overview}
% \end{figure}
\section{Overview}
To address the co-occurring issue, we propose a teacher-student framework that enforces cross-architecture consistency regularization to better separate foreground from background.~This enables us to derive a less biased classifier by leveraging the complementary strengths of two biased classifiers.~Our method can be interpreted as a regularization term that forces the classifiers to be cross-architecture consistency-aware and segmentation-aware.~The overall structure of the proposed framework is illustrated in Figure~\ref{fig:overview}.~We adopt multi-stage WSSS pipeline and only focus on the pseudo label refinement part.~Basically, the pipeline of our framework is to use knowledge transfer paradigm to train a segmentation-aware classification model using image-level labels first.~Then for each image,~extract the CAM to find regions relevant to smoke.~With original but coarse CAMs, advanced post-processing techniques can be applied to refine the pseudo masks.~Functionally, there are primary four components: a CNN network, a ViT network, knowledge transfer module and post-processing module. ~In the knowledge transfer module, we optimize two loss functions: a classification loss that is used for classification and a knowledge transfer loss that encourages the two networks to maintain consistency.
In the post-processing module, we combine both CRF and SAM-enhanced to improve the quality of pseudo labels.

The core motivation is that we find that for CAM generation, ResNet is good at spatial localization because of its local receptive field and weight sharing, while ViT shows superior boundary clarity due to its attention mechanism.~Thus, we aims to train a seed model that takes full advantage of ResNet and ViT while compensating their knowledge bias.~Unlike previous works relying on a single architecture model, our knowledge transfer scheme  enriches the classifier’s representation by providing more diverse internal knowledge without additional external knowledge or supervision, which can be seen as an internal supervision or regularization.

\section{CAM Refinement}
To obtain more accurate pseudo-mask, post-processing techniques like CRF or SAM refined methods can be applied to the initial CAMs to expand the coverage. 
However, post-processing often struggles with two common cases, wrongly activated pixels (False Positives) and misclassifications (CAM activation off-target).
Therefore, improving the classifier’s feature representation is crucial to reduce erroneous activations.~Thus,~Our proposed framework refines the quality of pseudo masks both during training and inference time.~For the model training,~our knowledge transfer method can be used to address co-occurrence issue via cross-architecture consistency regularization.
Once the co-occurring issue being tackled, the remaining challenges primarily lie in the use of fixed thresholding and partial object coverage, which can be easily tackled by post-processing methods.

\subsection{Optimizing feature representation}
Due to the single classification loss, CAMs fails to be segmentation aware, which suggests that extra constraints or regularization are needed.~One core observation from us is that if we only use the classification loss, the ViT model will wrongly activate background cues and the ResNet model will suffer from inaccurate boundary.~To handle this problem, we adopt ideas from knowledge distillation and consistency learning to regularize class activation map generation. Here we focus on exploring  cross-architecture consistency in feature level,~as feature map provides richer information than logits.~Another reason is that we want the ViT to match ResNet more in internal behavior, as better output does not mean better CAM.
% This is because the classifier learns to predict the correct class label very well (great train accuracy), but doesn’t learn where the object is located.
% \subsubsection{Cross-View Consistency }
% Before we talks about the details of our cross model consistency regularization.
% The main idea behind this method is that a good classifier is supposed to give invariant predictions under different view,thus we introduce consistency from different views to encourage the model focus more on the foreground object.\textbf{Augmented view consistency}:
% This strategy regularizes the model by aligning logits from the original image x and its augmented version.
% Specifically,the goal is to make the classifier make similar predictions for both version.
% It is worth mentioning that this method only  knowledge in the logit space while more reliable and informative knowledge does exist in the feature space.

% \textbf{global-local view consistency}
% As smoke sometimes occurs with small-sized, we want to enhance the model’s capability to detect these smoke.
% We aims to transfer the attention knowledge captured by the local views to the global network.

\subsubsection{Cross-Architecture Consistency Regularization Module }
\begin{figure}
    \centering
    \includegraphics[width=1\linewidth]{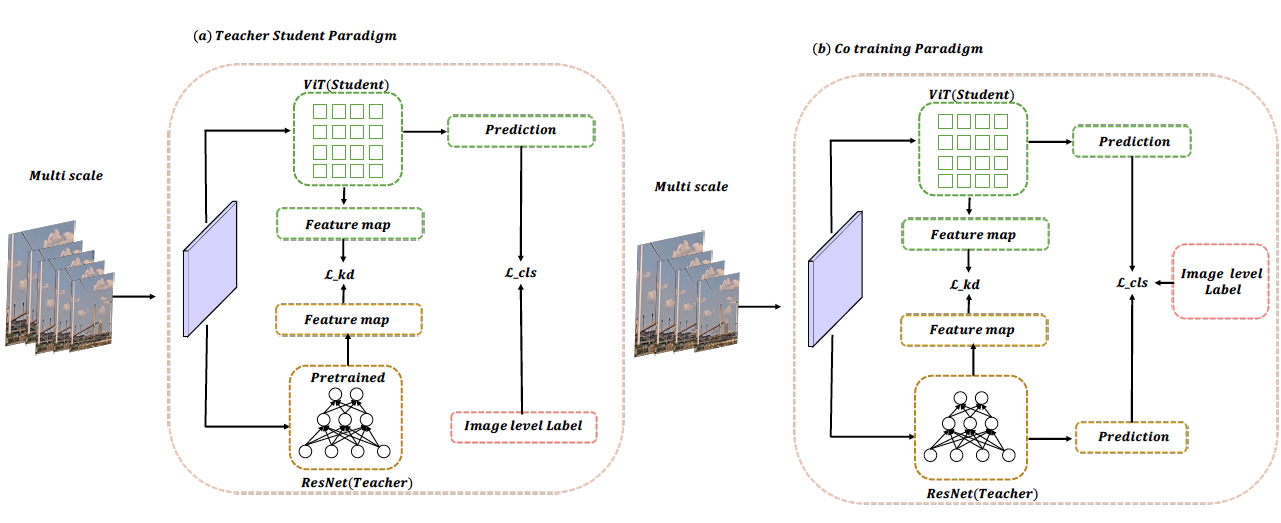}
    \caption{Different knowledge transfer paradigms}
    \label{fig:Paradigm}
\end{figure}

Cross-view consistency, which is widely adopted by previous works~\cite{wang2020self,jiang2022l2g}, only relies on a single model to seek better performance.
However, we argue that  a unitary network model has  limited learning capability, thus leading to incomplete representation in the feature space.~Although complementary information under multi-view enhances feature representation, its effectiveness remains constrained when confined to a single architecture.~Due to their inherent design, CNNs characterized by restricted receptive fields and weight sharing, tend to focus on local patterns and often overlook long-range dependencies.~In contrast, ViT capture the global dependencies of feature through self-attention.~Therefore, to leverage the complementary strengths of both architectures, we devise a knowledge transfer  loss function that enforces cross-architecture consistency between the feature representations of two distinct and independently initialized networks.
There are mainly two distinct knowledge transfer paradigms we can adopt, as shown in Figure~\ref{fig:Paradigm}.

\textbf{Teacher-Student Paradigm}
In this paradigm,~we adopt a teacher-student knowledge transfer framework, where a ResNet serves as the teacher model to guide a student model~(Vision Transformer) through unidirectional knowledge transfer. 
The parameters of the two networks are independent. Only the parameters of the student model are updated based on the classification and knowledge transfer loss. 
The teacher network is trained in advance and it keeps unchanged during training the student model.~In this setup,~ResNet’s well-localized feature maps help guide the ViT to reduce the contextual bias of co-occurrence, mitigating the wrongly activation often observed in Transformer-based representations.

\textbf{Co-training Paradigm}
In this framework,~two peer networks with distinct architecture~(ViT and ResNet) are trained and collaboratively guide each other in an online manner during the knowledge transfer process.~Both the teacher and student model are updated based on the classification and knowledge loss,~which can be regarded as a form of bidirectional feature distillation. 
% The typical teacher-student method is not suitable in this context.As both networks have their own advantages and limitations.
% For example,if we use ResNet as the teacher,ViT will  also inadvertently learn ResNet's over smoothing pattern,which will do harm to the overall performance. 
% Instead,we adopt a co-teaching strategy, where both networks learn collaboratively and complement each other. 

% This mutual learning process leverages the strengths of both architectures while mitigating their own weaknesses.

\paragraph{Knowledge Transfer Consistency Loss}
To preserve both spatial structures and semantic information and perform knowledge transfer effectively,~we devise a knowledge transfer consistency loss that encourages alignment between the feature representations of different architectures.
Basically,~there are mainly three different strategies for feature alignment:~spatial-wise,~channel-wise and global.~The global alignment approach focuses on capturing holistic representations, encouraging overall semantic similarity between the models.
The spatial-wise alignment emphasizes fine-grained correspondence by enforcing consistency at each spatial location, which is particularly beneficial for preserving local structural details.~The channel-wise alignment focuses on aligning feature distributions across channels, preserving semantic concepts encoded within each feature map dimension.
% Here we use channel-wise strategy to demonstrate how we measure feature consistency.
% Specifically, we compute a similarity matrix between each ViT channel and each ResNet channel to enforce channel-wise alignment. 

To create a unified feature map,~the critical step is the alignment of the features obtained from ResNet and ViT, which combines the distinctive knowledge captured  by both architectures. Inspired by~\cite{wahid2024hybrid}, we introduce a  learnable feature projection layer in both networks to match the student and teacher dimensions for better knowledge transfer of heterogeneous features. As mentioned by~\cite{miles2024understanding},~use a larger projector network does not necessarily improve the students performance, thus a linear projection layer is employed. This additional projection layer network's respective features into a shared channel space, ensuring channel-wise consistency and enabling more effective feature alignment.

\textbf{Feature Alignment}
Let \( \mathbf{F}^{(v)} \in \mathbb{R}^{C_{\text{v}} \times H \times W} \) be the ViT feature map~(patch tokens without class token) and
\( \mathbf{F}^{(r)} \in \mathbb{R}^{C_{\text{r}} \times H \times W} \)be the ResNet feature map.

\emph{Spatial Map}
Spatial Map is first map 3D feature map into 2D(HW), where the information of channel are averaged or maximized.

\emph{Inner product} Inner product is to use a matrix to convert channel-wise correlation with shape BCC.

\emph{Feature projection}
Both of  ViT and ResNet model include a linear projection to build a channel-wise feature alignment between them.~Subsequently,~after feature projection,~we can  calculate the consistency between the aligned feature map more precisely.~Given the ViT feature map \( \mathbf{F}^{(v)} \in \mathbb{R}^{C_{\text{v}} \times H \times W} \), where \( C_{\text{v}} \) is the embedding dimension and \( H \times W \) denotes the spatial resolution, we apply a 1{\texttimes}1 convolutional layer to project it into a 2-channel space:

\begin{equation}
\tilde{\mathbf{F}}^{(v)}(x, y) = \mathbf{W}_{\text{p}} \cdot \mathbf{F}^{(v)}(x, y) + \mathbf{b}_{\text{p}}, \quad \tilde{\mathbf{F}}^{(v)} \in \mathbb{R}^{2 \times H \times W}
\label{eq:projection}
\end{equation}

Here, \( \mathbf{W}_{\text{proj}} \in \mathbb{R}^{2 \times C_{\text{v}}} \)  and \( \mathbf{b}_{\text{proj}} \in \mathbb{R}^{2} \) are the learnable weights matrix and bias of the feature projection layer, respectively. The transformation is applied at every spatial location, ensuring consistent projection across the entire feature map.

\textbf{Calculate Consistency}
For computing similarity,~one common way is to  first convert the activation of the feature map into a probability distribution such that the discrepancy can be measured using a probability distance metric such as the KL divergence.~However,~this way to handle and extract knowledge may collapse a lot of information.
Instead,~here we opt to directly minimize the differences between the raw feature maps from teacher and student networks using  similarity metrics like cosine similarity.~Both feature maps are flattened across spatial dimensions.~The global cosine similarity between the feature representations of ViT and ResNet is defined as:

\begin{equation}
S_{\text{global}} = \frac{\mathbf{f}^{(v)} \cdot \mathbf{f}^{(r)}}{\|\mathbf{f}^{(v)}\| \cdot \|\mathbf{f}^{(r)}\|}
\label{eq:global_similarity}
\end{equation}

where $ \mathbf{f}^{(v)} \in \mathbb{R}^{C_v} $ and $ \mathbf{f}^{(r)} \in \mathbb{R}^{C_r} $ are the global average pooled feature vectors from ViT and ResNet respectively.

The knowledge transfer loss based on global similarity is then formulated as:
\begin{equation}
\mathcal{L}_{\text{global}} = 1 - S_{\text{global}}
\label{eq:global_transfer_loss}
\end{equation}

Our final loss is the combination of the binary cross entropy loss and knowledge transfer loss:
\begin{equation}
\mathcal{L} = \mathcal{L}_{\text{cls}} +\lambda\mathcal{L}_{\text{global}} 
\label{eq:loss}
\end{equation}

where, $\lambda$ is the coefficient that controls the trade-off between the two loss terms.

\subsection{Post-processing}
\begin{figure}[H]
    \centering
    \includegraphics[width=0.55\linewidth]{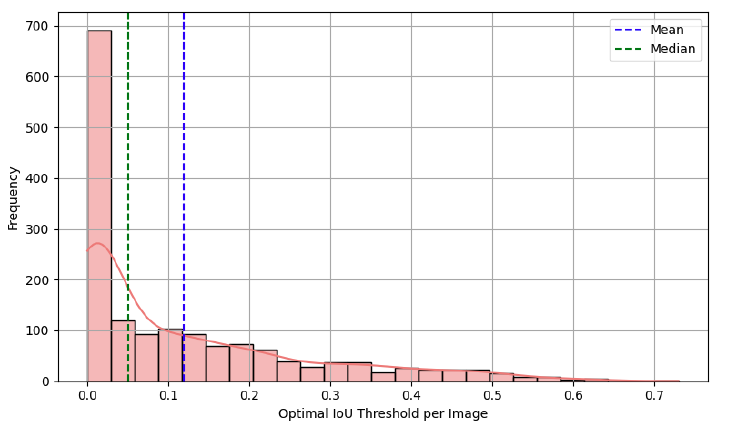}
    \caption{Distribution of Optimal Iou Thresholds on Testing Set.}
    \label{fig:threshold_vit_s}
\end{figure}

Due to CAM ambiguity,~the generated pseudo masks are often incomplete and suffer from low activation responses.~Post-processing techniques are mainly used to solve the partial  and low activation  issue of CAMs at the inference time, which can also be regarded as object expansion technologies.~From Figure~\ref{fig:threshold_vit_s} we can see that the optimal threshold value for converting CAMs to pseudo masks varies across image.~The common practice is to apply a fixed threshold for all images.~As adjusting the threshold for each image or dataset individually is impractical and difficult to scale,~post-processing serves as a general and effective solution to refine pseudo masks.

% \subsubsection{Sliding windows}
% utilize the sliding window strategy during inference and aggregate the attention maps from different image patches

% Random walk is applied by most WSSS algorithm to refine the CAM\cite{li2021pseudo}.It requires CRF operations at different $\alpha$ values to synthesize the training data.
\subsubsection{SAM-enhanced}
This post-processing method is first proposed in~\cite{chen2023segment}.~SAM is very powerful and tends to provide masks of all the distinguishable instances in the image.~In our framework,~we leverage SAM’s boundary-aware capabilities to enhance CAMs, where SAM is kept frozen, serving only for inference purposes.
In default,~we provide  prompt encoder with sparse or dense prompts. However,~in our case, the segmentation object (smoke) is determined, so the input of the prompt is simplified to none. ~We first generate CAMs and use CAMs as the initial seed, which tell SAM the exact mask that we need. Then generate SAM masks which return a list of object masks.~And SAM masks with enough overlap~($\geq0.3$~IOU) are selected as foreground masks.~Then,~several strategies can be employed to fuse the initial seed with CAMs generated by SAM, including AND fusion, OR fusion, and direct copy.~Therefore, the final result is masks that is semantically correct~(guaranteed by CAMs) and spatially precise~(guaranteed by SAM).
% with the CAM-derived pseudo-labels are retained Meanwhile,SAM masks that substantially
% encompass the CAM-derived pseudo-labels can also be collected, targeting the challenge of partial activation.
% The default embedding is a learnable vector that is added to each position of the image embedding.
% Transfers the knowledge of SAM to the classifier during the training process, enhancing the quality of CAMs itself.

\subsubsection{CAM Fusion}
\begin{figure}[H]
    \centering
    \includegraphics[width=0.75\linewidth]{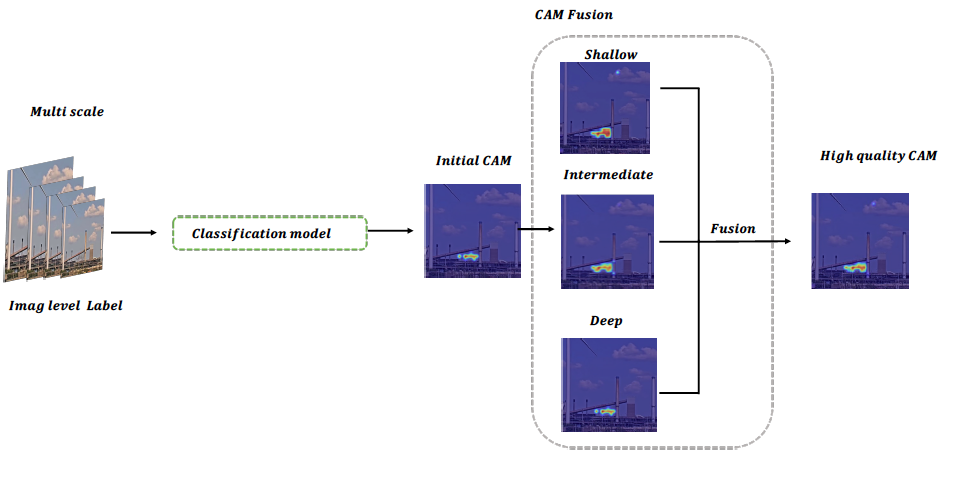}
    \caption{CAM Fusion}
    \label{fig:CAM_fusion_view}
\end{figure}

Due to the low spatial resolution of the final convolutional layer, standard Class Activation Maps tend to highlight only coarse regions of the target objects. However, CAMs extracted from different layers are often complementary~\cite{jiang2021layercam}. This is based on the fact that shallower layers contribute precise spatial localization, whereas deeper layers provide semantic abstraction.~Motivated by the fact that intermediate layers preserve fine-grained details while deeper layers capture high-level semantic information, instead of relying solely on the final layer for CAM generation, we select multiple intermediate blocks as shown in Figure~\ref{fig:CAM_fusion_view} to do feature fusion.~In addition,~the final layer, while highly abstract, is also more prone to errors due to overgeneralization or misinterpretation of context.~In contrast, shallower layers are less abstract and more directly represent visual content, reducing the likelihood of incorrect activations.

% \subsubsection{text driven supervision}
% Contrastive Language-Image Pre-training(CLIP) models include two architectures, e.g., ResNet-based and ViT-based. Note that Grad-CAM is not only applicable to CNN-based architecture but also works on the vision transformer. In this paper, we leverage the ViT-based CLIP model because the CNN-based model fails to explore the global context and suffers from the discriminative part domain heavily. 

\subsubsection{Conditional Random Field}
Conditional random field~(CRF)~\cite{chen2017deeplab}  is a graphical model used for refining CAM to pseudo masks by using color and position information as features~\cite{shimoda2019self}.
CRF tries to smooth and regularize the input probability maps using image cues, which can suppress low-activation regions.~Unary and pairwise potentials are two main components of CRF model. Unary potentials refer to the confidence of assigning the class label to a single pixel  based on initial predictions.~Pairwise potentials defined by a linear Gaussian kernels,~works on all pairs of pixels in the images.~However,~CRF does not guarantee any improvement in the mean IoU score, and it sometimes degrades the segmentation masks~\cite{shimoda2019self}.

\subsubsection{CLIP-Aided}
Most recently,~using external natural language supervision to refine pseudo labels has emerged as a promising trend~\cite{xie2022clims}.~Contrastive Language-Image Pre-training~(CLIP)~\cite{radford2021learning},~known for its strong zero-shot performance,~is trained on a large-scale of text paired with images found across the internet. 
Following this methodology,~we explore  CLIP-aided post-processing approach to improve the quality of pseudo labels.~Intuitively, combining the natural language supervision will give us better pseudo masks 
as describing smoke in natural language do provide a clear guidance than image-level labels.~For example, textual descriptions of "smoke" are expected to  provide clearer and more consistent semantic cues for foreground localization task.

\begin{figure}[H]
    \centering
    \includegraphics[width=0.7\linewidth]{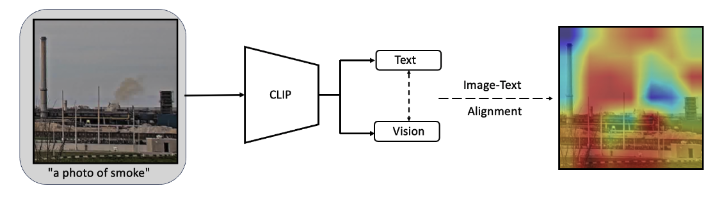}
    \caption{Use CLIP to generate CAMs.}
    \label{fig:enter-label}
\end{figure}

\subsubsection{Multi scale input}
Given that single-scale image inputs often fail to capture features effectively across varying object sizes and spatial contexts, leveraging multi-scale features has become a widely adopted strategy~\cite{chen2016attention}. The underlying reason is that the importance of  spatial  features may vary depending on the resolution of images.~One common approach to obtain multi-scale features is to feed multiple resized versions of the same input image into the model,~which is adopted in our work.

\subsubsection{Affinity with Random Walk}
Affinity based post-processing technique is first proposed in~\cite{ahn2018learningpixellevelsemanticaffinity}.~This kind of method models  the pixel-level affinity distance from initial CAMs using local RGB and position information and applied a random walk to propagate individual class labels at the pixel level without extra knowledge.

% \subsection{Contrastive Loss}
% Contrastive loss regularizes feature learning,leading to smoother segmentation regions and sharper object boundaries.
% Formally,we define the distance between two data points as
% $$ D_{ij}= _{{aff}} \tag{1.1} $$
% The constrastive loss and triplet loss.The SEC loss and CRF loss

% \begin{equation}\label{total_loss}
%  \mathcal{L}_{\text{total}} = \mathcal{L}_{cls}+\mathcal{L}_{seg}+
%  \lambda \mathcal{L}_{cont} +
% \end{equation}
% Doing contrastive learning on the pixel level.

% \subsubsection{Info Noise-Contrastive Estimation(InfoNCE)}
% exponential moving average (EMA),
% \begin{equation}\label{ema}
% \theta^{ema}_t=\lambda \theta^{ema}_{t-1}+(1-\lambda)\theta_t
% \end{equation}

% \begin{equation}\label{eq:contrastive}
% \mathcal{L}_{InfoNCE}=w^T_cf^{cam}(x,y)
% \end{equation}

% \section{Robust learning given noisy pseudo-labels}
% The noise in pseudo-masks in inevitable and impedes the training of segmentation model.

% The pseudo-labels with a higher mIoU do not mean a better segmentation model.
% As a corollary,

% Only a few works attempt to suppress the noise at this stage.

%\input{content/chapter-2}
%\input{content/chapter-3}
%\input{mainmatter/chapter-4} % Create file to add

\chapter{Evaluation}
\label{chapter:result}
In this section,~we evaluate our approach with extensive experiments.

\begin{figure}[H]
    \centering
    \includegraphics[width=0.8\linewidth]{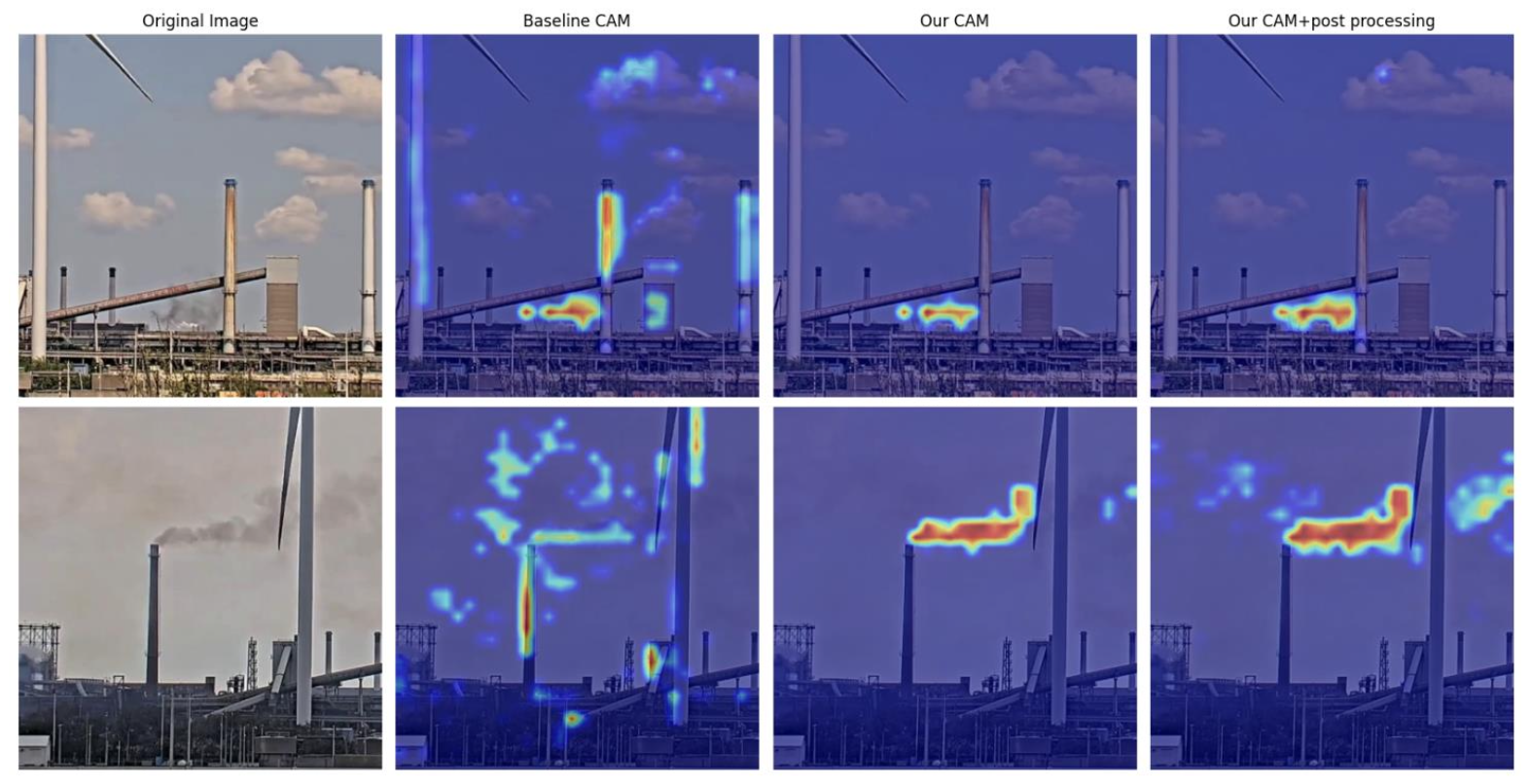}
    \caption{Comparison of CAMs generated from our method and baseline. Our method accurately highlights the foreground regions without co-occurrence issue.}
    \label{fig:baselinevs}
\end{figure}

\section{Experimental setup}
We design a series of experiments to evaluate the effectiveness of our proposed framework in various configurations.~Note that for our WSSS framework,~only image-level classification labels are available during network training.~Specifically, we train and evaluate the following models:
\begin{itemize}
\item Baseline CNN: A standard ResNet-based classifier trained solely on training dataset with image-level labels.~This setup is used to assess the standalone performance of ResNet in weak supervision.
\item Baseline ViT~(1): A Vision Transformer classifier trained on the training set with image-level labels only. This setup is used to assess the standalone performance of ViT in weak supervision.
\item Baseline ViT~(2): A Vision Transformer classifier trained on the combined training set and a portion of the test set, using only image-level labels. This setting simulates a scenario with slightly more data access but no additional supervision.
\item Baseline ViT~(3): A Vision Transformer classifier trained on both the training set and a subset of the test set, using a mixture of image-level and pixel-level labels. 
\item ViT + CNN Co-training: A hybrid framework in which ResNet and ViT are jointly trained on the training set with image-level labels. This setup is used as a baseline to compare against our teacher-student paradigm.
\item \textbf{Teacher-Student Paradigm~(Ours)}: We implement a teacher-student framework, where a pretrained ResNet-50 model serves as the teacher and a ViT-B acts as the student, incorporating a cross-architecture consistency regularization to enhance feature alignment and localization performance.
\item Fully supervised model for comparison: We additionally train fully supervised segmentation counterparts, such as fine-tuned versions of SAM, to serve as upper-bound references for evaluating the performance gap between weakly and fully supervised approaches.
\end{itemize}

To examine the contribution of each component, we conduct ablation studies by selectively enabling/disabling:
\begin{itemize}
\item Cross-architecture knowledge transfer consistency loss.
\item Different variants for feature alignment and computing consistency loss.
\item Advanced post-processing techniques such as SAM-enhanced, and CAM fusion to improve pseudo-mask quality.
\end{itemize}

\textbf{Dataset and evaluation metrics}
We conduct our experiments on our custom dataset.
Our dataset consists of both images  and videos depicting industrial smoke and non-smoke scenes.The videos were captured in the IJmond region of Amsterdam, the Netherlands, across several specific geographic locations. ~There are in total 481 videos,~each comprising  36 frames,~annotated by experts or trained volunteers using the AI tool~(IJmondCAM) \footnote{\url{https://ijmondcam.multix.io/}} developed by us. All video frames were extracted, and about 2400 frames are manually verified by us to ensure annotation consistency.~The training dataset comprises a total of 2,488 images,~with the resolution of 900×900. These images were manually labeled into two categories:  Smoke (1,556) and Non-Smoke (932). 
The test dataset consists of 900 high-resolution images~(1920×1080) with pixel-level annotations provided by experts.~These are normalized and cropped into 512×512 patches using a sliding window approach.~Only patches contains visible smoke will be used for evaluation.
However,~our dataset has relatively low variability in background contexts, posing a challenge for weakly supervised learning,~as models could potentially misinterpret noise or co-occurring objects as relevant features. Therefore, to improve diversity,~we also apply supplementary images as out-of-distribution data for training,~which comes from The Rise~\cite{hsu2021project} and Smoke5K~\cite{yan2022transmission}.~By default, mean Intersection-over-Union~(mIoU) is used as the evaluation criteria.~Since this is a binary classification task, the reported mIoU corresponds specifically to the smoke class.~To evaluate the quality of pseudo masks,~we generate them for every image in the test set and then use the ground truth to compute mIoU.
% that contain co-occurred background objects to facilitate the discrimination capability of the network. 

\section{Implementation details}
For the fair comparison,~we conduct all experiments on a single NVIDIA A100 GPU and implement our framework using PyTorch.

\textbf{Classification Network}
During the training stage,~we use the ResNet50 and ViT-base (ViT-B) as the backbone. The feature backbone is initialized with ImageNet~\cite{deng2009imagenet} classifier weights.~The remain components are then randomly initialized and trained from scratch.~For classification,~we adopt binary cross-entropy loss. Besides, we also employ a pixel correlation module (PCM)~\cite{wang2020self} into the ResNet classification network to constrain the shape of the target object for ResNet.~To guarantee the backbone accepts input images of arbitrary size, the pos embedding of ViT will be resized to input size via bilinear interpolation.~For our proposed classifier,~we train the networks for 3 epochs with a batch size of 8.~We use AdamW optimizer instead of Adam and SGD.~We also tried the widely used backbone ResNet38d~\cite{wu2019wider} and mobileNetv2~\cite{sandler2018mobilenetv2}.
However, these models perform poorly  in our setting.~To determine the appropriate learning rate, we performed a learning rate range test and choose the learning rate for the classifier to be $1e-4$.

\textbf{Segmentation Network}
For comparison, we also implement  fully supervised semantic segmentation pipeline using SAM fine-tuning and SERT~\cite{strudel2021segmenter}. For SAM fine tuning, we use partial fine-tuning strategy by freezing the image encoder and prompt encoder, training  only the mask encoder and a learnable  task embedding.

\textbf{{Training Parameters and Inference Strategy.}}
% For cross view consistency learning, we avoid transformations that alter the spatial layout, as such changes can degrade overall performance. Instead, we apply only weak augmentations, such as color perturbations that adjust the brightness, contrast, saturation, and hue of the images. To enforce consistency, we directly compute the Mean Squared Error (MSE) loss between the raw logits produced by the model for the original and augmented images.
In the context of WSSS training, we observe that image-level augmentations do not provide significant benefits for our setting. Therefore, we do not apply any data augmentation during this stage.
During inference, it is common practice to use multi-scale testing, which aggregates predictions from multiple image scales to improve final performance. To assess the effectiveness of our method, we evaluate CAMs generated under both single-scale and multi-scale settings. For multi-scale testing, we use scale ratios of {0.5, 1.0, 1.5, 2.0}.~For thresholding, if a pixel’s maximum activation score is below a fixed low threshold (set to 0.3 in our experiments), it is assigned to the background class.

\section{Experimental Results}
\begin{figure}[H]
    \centering
    \includegraphics[width=0.9\linewidth]{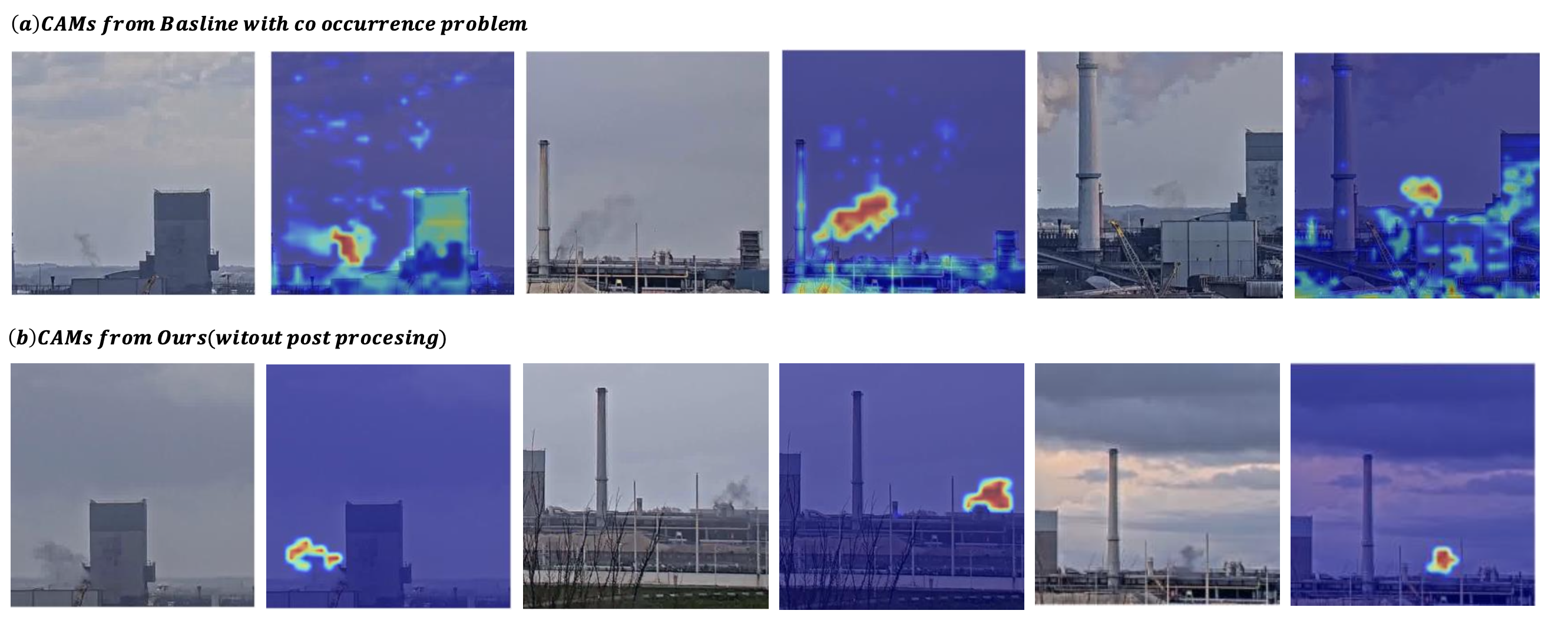}
    \caption{Comparison of CAMs generated from our method and baseline.}
    \label{fig:cooccurrence}
\end{figure}

We experimentally verify the effectiveness of the proposed framework.~In order to clearly compare the effect,we remove post-processing for comparison.
Our generated CAMs are more accurate to match the ground truth than the baseline. ~As shown in Figure~\ref{fig:cooccurrence}, the CAMs generated by our method can correctly locate the foreground with accurate boundary while shows no co-occurring patterns.

\textbf{Training:}
First, we compare our final result with baseline,~indicting substantial improvement in the quality~(mIoU) of generated seed masks as shown in Figure~\ref{tab:res_vit}.~The model trained using our approach achieves significant improvement.~Additionally, we also trained the ViT backbone mixed with part of the test data,~which boosts the performance and achieves mIOU 47.99.~Furthermore, we also examine using weak image-level annotations in addition to limited pixel-level annotations.
% Additionally, we observe that the cam generated by ResNet architecture  fails determine accurate object boundaries, while Transformer based network does much better as shown in Fig \ref{tab:res_vit}.
% We also test if we  have  limited ground truth masks to guide the classifier,then if the performance of the classifier will improve.

\begin{table}[ht]
\centering
\caption{Evaluate mIOU of pseudo masks with different backbones.~$T_1$:Train dataset.~$T_2$:Test dataset.
For pixel-level labels,~we simulate the real world setting that only a fraction of pixel-level annotations available.~Gray rows indicate ours method.}
%adamW  lr=1e-4  epoch =10
\label{tab:res_vit}
\vspace{-0.5em}
\begin{tabular}{lccc}
\toprule
\textbf{Supervision}& \textbf{Source} & \textbf{Backbone} & \textbf{mIOU} \\
\midrule
image-level & $T_1$&ResNet50& 26.10\\
image-level & $T_1$& ResNet101& 21.29\\
image-level &$T_1$ & ViT-S & 13.18\\
\rowcolor{gray!20}
image-level &$T_1$ & Ours & 47.37\\
% image-level &$T_1$ & ViT-B & \\
image-level &$T_1+T_2$& ViT-B & 47.99\\
image-level+limited pixel-level&$T_1$& ViT-B &  \ding{55}\\
\bottomrule
\end{tabular}
\end{table}

\textbf{Comparison to fully supervised counterparts}
We present the comparison of the performances of our WSSS approaches with several different networks trained in fully supervised learning manner in Table~\ref{tab:fully}.

\begin{table}[H]
\centering
\caption{Comparison of semantic segmentation methods.~Performance of fully supervised learning methods are trained with ground truth labels without any post-processing.}
\label{tab:fully}
\vspace{-0.8em}
\begin{tabular}{llc}
\toprule
\textbf{Method}  & \textbf{Backbone} & \textbf{mIOU} \\
\midrule
\multicolumn{3}{l}{\textit{Fully Supervised}} \\
\midrule
%epoch 10
% DeepLabV3+\textsuperscript{\cite{chen2018encoderdecoderatrousseparableconvolution}} & ResNet101 & \\
% SegFormer\textsuperscript{\cite{}} & MiT&  \\
% %epoch 100
SERT\textsuperscript{\cite{xie2021segformer}} &Transformer & 68.27 \\
SAM-fine-tuning & ViT-B & 54.68 \\
\midrule
\multicolumn{3}{l}{\textit{Multi stage WSSS}} \\
\midrule
Ours+post-processing &ViT-S+ResNet50&52.93 \\
\bottomrule
\end{tabular}
\end{table}

\textbf{Comparison with several pseudo masks refinement methods}
We demonstrate the effectiveness of the proposed scheme by comparing it with several previous methods.
During our experiment,~we find that our method performs  better than other methods as presented in Table~\ref{tab:previous}.
% Our framework outperforms previous methods by a large margin on initial seeds. CRF could further boosts the performance ,which even outperforms previous methods with extra affinity networks. The result is accurate enough, hence the stage of training an affinity network is omitted.

\begin{table}[ht]
\centering
\caption{Comparison with previous methods.}
\label{tab:previous}
\vspace{-0.5em}
\begin{tabular}{llc}
\toprule
\textbf{Method} & \textbf{backbone} &  \textbf{mIOU}\\
\midrule
TransCAM \textsuperscript{\cite{li2023transcam}} & Conformer &15.02\\
AffinityNet\textsuperscript{\cite{ahn2018learningpixellevelsemanticaffinity}}   & ResNet50 & 24.28\\
% Cross-view consistency loss &ResNet101&29.73\\
% ToCo &   & {--} &\\
PCM\textsuperscript{\cite{wang2020self}} &ResNet50&33.56\\
\midrule  % 
\rowcolor{gray!20}
Ours &ViT-S+ResNet50  & 47.37\\ 
Ours+post-processing &ViT-S+ResNet50  &52.93 \\ 
\bottomrule
\end{tabular}
\end{table}

% \begin{tabular}{cccS[table-format=2.2]}

%     \toprule
%     $L_{\text{cls}}$ & $L_{p\text{-}cls}$ & $L_{re}$ & {mIoU (\%)} \\
%     \midrule
%     \checkmark &  &  & 47.82 \\
%     \checkmark & \checkmark &  & 47.70 \\
%     \checkmark &  & \checkmark & 49.21 \\
%     \checkmark & \checkmark & \checkmark & 51.53 \\
%     \bottomrule
% \end{tabular}

\textbf{Post-processing}
% As previously discussed, CAMs tend to highlight only small, highly discriminative regions of the object.To address this limitation, various post-processing techniques are applied to expand CAM coverage.
Table~\ref{tab:Post-processing} displays the performance of different post-processing techniques.
We apply these post-processing techniques to both the  strongest variant and a weaker one of our method across different settings to evaluate their robustness.~This analysis allows us to assess not only the general effectiveness of the post-processing itself but also the resilience of our models. It is worth mentioning that if the best-performing version of our model achieves significant performance gains with minimal reliance on post-processing~(from cam fusion we can observe this pattern), it indicates a higher degree of inherent robustness and a reduced dependency on external refinement strategies.~Some visualizations are available in Appendix.

\begin{table}[H]
\centering
\caption{Evaluation of pseudo labels with different post-processing techniques. \ding{55} means that this method failed in our task.}
\label{tab:Post-processing}

\begin{subtable}[t]{0.48\textwidth}
\centering
\begin{tabular}{lcc}
\toprule
\textbf{Method} & \textbf{mIoU} & \\
\midrule
w/o post-processsing &  37.42&\\
+ Multi scale &   38.49 &\\
+AffinityNet\textsuperscript{\cite{ahn2018learningpixellevelsemanticaffinity}}   &  34.00 &\\
+SAM-enhanced\textsuperscript{\cite{kirillov2023segment}} & 43.20 &\\
+CLIP\textsuperscript{\cite{radford2021learning}}& \ding{55}  &\\
+CRF& 43.27  &\\
+CAM fusion&  46.91 &\\
+CRF+CAM fusion& 37.81  &\\
+CRF+AffinityNet\textsuperscript{\cite{ahn2018learningpixellevelsemanticaffinity}}   &  38.51 &\\
Optimal threshold & 53.92&\\
\bottomrule
\end{tabular}
\caption{CAMs generated by ours (Worse seed)}
\label{subtab:postprocessing1}
\end{subtable}
\hfill
\begin{subtable}[t]{0.48\textwidth}
\centering
\begin{tabular}{lcc}
\toprule
\textbf{Method} & \textbf{mIoU} & \\
\midrule
w/o post-processsing &  46.25&\\
+ Multi scale &   47.37 &\\
+CAM fusion&  45.27 &\\
+CRF+AffinityNet\textsuperscript{\cite{ahn2018learningpixellevelsemanticaffinity}}   & 49.16  &\\
+SAM-enhanced\textsuperscript{\cite{kirillov2023segment}} &  51.00&\\
+CLIP\textsuperscript{\cite{radford2021learning}}& \ding{55}  &\\
+CRF&  52.52 &\\
+CRF+SAM-enhanced&  52.93 &\\
Optimal threshold &57.15 &\\
\bottomrule
\end{tabular}
\caption{CAMs generated by ours (Best seed)}
\label{subtab:postprocessing2}
\end{subtable}
\end{table}

For cam fusion, significant improvements are observed when the initial seed is of low quality.~However, when the initial seed is already strong, this method does harm to the quality of pseudo masks.
\begin{table}[htbp]
\centering
\caption{Impact of cam fusion.}
\label{tab:cam_impact}

% 第一个子表
\begin{subtable}[t]{0.48\textwidth}
\centering
\begin{tabular}{lcc}
\hline
\textbf{Layer} & \textbf{Seed} & \textbf{mIOU} \\
\hline
\rowcolor{gray!20}
-2,-1 & 47.37 &  45.27\\
-5,-1 & 47.37 &  40.82\\
-5,-4,-2,-1 & 47.37 &  35.34\\
\hline
\end{tabular}
\caption{Layer chosen for fusion(Best seed)}
\label{subtab:size1}
\end{subtable}
\hfill
% 第二个子表
\begin{subtable}[t]{0.48\textwidth}
\centering
\begin{tabular}{lcc}
\hline
\textbf{Layer} & \textbf{Seed} & \textbf{mIOU} \\
\hline
\rowcolor{gray!20}
-5,-4,-2 & 38.49 & 46.91 \\
-4,-2    & 38.49 &  45.99\\
-4,-2,-1 & 38.49 &  45.11\\
-4,-1    & 38.49 &  46.90\\
\hline
\end{tabular}
\caption{Layer chosen for fusion(Worse seed)}
\label{subtab:size2}
\end{subtable}

\end{table}

The resolution of the input image really matters as shown in Table~\ref{tab:multi-scale}. We observe that mIoU varies significantly with different input scales. Moreover, leveraging multi-scale inputs allows our model to achieve the highest mIoU of 47.37.
\begin{table}[H]
\centering
\caption{Experiments with multi scale input images}
\label{tab:multi-scale}
\vspace{-0.5em}
\begin{tabular}{lcc}
\toprule
\textbf{Scale} & \textbf{Method} & \textbf{mIOU} \\
\midrule
0.5 & Ours & 40.17\\
1.0 & Ours & 46.25\\
1.5 & Ours  & 45.73\\
2.0 & Ours  & 42.60\\
\midrule 
(0.5,1.0,1.5,2.0)& ViT& 47.37\\ 
\bottomrule
\end{tabular}
\end{table}

\section{Ablation Studies}
To study the effect of each component of our method,~we make ablation studies in different settings.

\emph{1)Effectiveness of our framework and cross-model knowledge transfer consistency loss}:
We evaluate our method under various training settings to assess its effectiveness.
As we can see from the Table \ref{tab:paradigm_comparison},~teacher-student paradigm with cosine similarity and global feature alignment achieves the best performance.
We compare different knowledge transfer paradigms and observe that the teacher-student framework is more suitable for our task.
Moreover,~we conduct ablation studies comparing channel-wise,~global,~and spatial feature transfer mechanisms. 
As shown in Table \ref{tab:paradigm_comparison}, all three methods works, showcasing the effectiveness of our proposed knowledge transfer consistency loss.
Specifically,~the global feature alignment yields the most significant improvement, outperforming both the spatial-wise and channel-wise methods.
In addition, we apply various similarity metrics to calculate feature consistency.
As reported in Table \ref{tab:paradigm_comparison}, the cosine similarity demonstrates  superior performance compared to the other strategies and is used in our method.~As reported in Table~\ref{tab:projector},~we observe that the linear learnable projector layer can provide significant improvements in mIOU.

\begin{table}[htbp]
\centering
\caption{Comparison of different knowledge transfer strategies(+feature projection), including channel-wise versus global/spatial distillation, various similarity metrics, and distinct training paradigms.}
\label{tab:paradigm_comparison}
\begin{tabular}{cccccc}
\hline
\textbf{Paradigm} & \textbf{Teacher} & \textbf{Student} &\textbf{Metric} &\textbf{Level}&\textbf{mIOU} \\
\hline
\rowcolor{gray!20}
Teacher-Student & ResNet(Pre-trained) & ViT &Cosine& Global&47.37 \\
Teacher-Student & ResNet(Pre-trained) & ViT &$L_1$& Global&38.39\\
Teacher-Student & ResNet(Pre-trained) & ViT &$L_2$& Global&33.74\\
Teacher-Student & ResNet(Pre-trained) & ViT &Cosine& Spatial&43.70 \\
Teacher-Student & ResNet(Pre-trained) & ViT &Cosine& Channel&46.85\\
\midrule
Co-training& ViT + ResNet(From scratch) & ViT+ResNet&Cosine& Global&45.93 \\
Co-training& ViT + ResNet(From scratch) & ViT+ResNet&$L_1$&  Global&18.51\\
Co-training& ViT + ResNet(From scratch) & ViT+ResNet&$L_2$& Global&0.27 \\
Co-training& ViT + ResNet(From scratch) & ViT+ResNet&Cosine& Spatial&45.11  \\
Co-training& ViT + ResNet(From scratch) & ViT+ResNet&Cosine& Channel&42.75 \\
\hline
\end{tabular}
\end{table}

\begin{table}[htbp]
\centering
\caption{Impact of feature projection.}
\label{tab:projector}

\begin{tabular}{ccccc}
\hline
\textbf{Feature projection} & \textbf{Feature Alignment Level} & \textbf{Paradigm}&\textbf{Seed} & \textbf{mIOU} \\
\hline
\ding{51} & Global &Teacher-Student& 33.56 & 47.37\\
\ding{51} & Global &Co-training& 33.56& 45.93\\
\ding{51} & Channel &Co-training& 33.56 &  42.75 \\
\ding{55} & Channel &Co-training &33.56 & 38.49 \\
\hline
\end{tabular}

\vspace{0.2cm}
\footnotesize
\end{table}

\begin{table}[h]
\centering
\caption{Comparison of Feature-based and Logits-based Alignment Methods}
\label{tab:feature_vs_logits}
\begin{tabular}{lccccc}
\hline
\textbf{Alignment} & \textbf{Type} & \textbf{Metric} & \textbf{Paradigm} & \textbf{Seed} & \textbf{Refined} \\
\hline
Feature projection                    & Feature-based & Cosine       & T-S & 33.56             & 47.37             \\
Inner product + Feature proj   & Feature-based & Cosine       & T-S & 33.56             & 33.87             \\
Spatial map + Feature proj     & Feature-based & Cosine       & T-S & 33.56              & 48.19                 \\
Spatial map + Feature proj + norm.  & Feature-based & Cosine       & T-S &  33.56             &43.70                \\
\hline
Logits-level alignment   & Logits-based  & KL Divergence & T-S & 33.56                & 44.65               \\
\hline
\end{tabular}
\vspace{0.2cm}
\footnotesize
\end{table}

\emph{2) The effect of the transfer loss coefficient $\lambda$}: To explore the impact of $\lambda$, we increase the weight of transfer loss from 0.3 to 1.3.
Table~\ref{tab:consistency_loss} displays the mIOU of various coefficient values on our knowledge transfer loss function.~The results reveal that the cross model knowledge transfer loss shows  consistent superiority in mIoU  compared to the baseline without this term.
% However, when coefficient $\lambda$ reaches 1.2, the network's performance in worse setting drops, suggesting that feature alignment and projection plays a critical role in our knowledge transfer loss. 
% Without appropriate feature alignment and projection, when the transfer loss dominates,the model may confused with the messy heterogeneous feature representation and performs worse.
% suggesting that transfer loss dominates,suppressing classification.

\begin{table}[htbp]
\centering
\caption{The Effect of the Consistency Loss Coefficient $\lambda$. We choose the max mIOU from baseline model (ResNet50+PCM) for comparison.}
\label{tab:consistency_loss}

\begin{tabular}{cc}
\toprule
\textbf{$\lambda$} & \textbf{mIOU}  \\
\midrule
0.3 &  38.55\,$\textcolor{blue}{\uparrow 4.99}$  \\
0.5 &  43.81\,$\textcolor{blue}{\uparrow 10.25}$  \\
0.8 & 45.04 \,$\textcolor{blue}{\uparrow 11.48 }$  \\
\rowcolor{gray!20}
1.0 & 47.37\,$\textcolor{blue}{\uparrow 13.81}$  \\
1.3 &  46.62\,$\textcolor{blue}{\uparrow 13.06}$  \\
1.5 &  41.90\,$\textcolor{blue}{\uparrow 8.34}$  \\

\bottomrule
\end{tabular}
\end{table}

% \begin{table}[htbp]
% \centering
% \caption{The Effect of the Consistency Loss Coefficient $\lambda$. We choose the max mIOU from baseline model (ResNet50) for comparison.}
% \label{tab:consistency_loss}

% \begin{subtable}{0.48\textwidth}
% \centering
% \begin{tabular}{cc}
% \toprule
% \textbf{$\lambda$} & \textbf{mIOU}  \\
% \midrule
% 0.8 & 45.04 \,$\textcolor{blue}{\uparrow18.94 }$  \\
% \rowcolor{gray!20}
% 1.0 & 47.37\,$\textcolor{blue}{\uparrow 21.27}$  \\
% 1.3 &  46.62\,$\textcolor{blue}{\uparrow 20.52}$  \\
% \bottomrule
% \end{tabular}
% \caption{CAMs generated by ours (Best setting)}
% \label{subtab:lambda2}
% \end{subtable}

% \end{table}

% \begin{subtable}{0.48\textwidth}
% \centering
% \begin{tabular}{cc}
% \toprule
% \textbf{$\lambda$} & \textbf{mIOU}  \\
% \midrule
% 0.1 & 31.30 \,$\textcolor{blue}{\uparrow 5.2}$   \\
% 0.2 & 32.24 \,$\textcolor{blue}{\uparrow 6.14}$   \\
% 0.5 & 36.74 \,$\textcolor{blue}{\uparrow 10.64}$  \\
% 0.6 & 32.60 \,$\textcolor{blue}{\uparrow 6.5}$   \\
% \rowcolor{gray!20}
% 1.0 & 38.49 \,$\textcolor{blue}{\uparrow 12.39}$  \\
% 1.2 & 24.60 \,$\textcolor{red}{\downarrow 1.5}$  \\
% \bottomrule
% \end{tabular}
% \caption{CAMs generated by ours(Worse setting)}
% \label{subtab:lambda1}
% \end{subtable}
% \hfill
% 第二个子表

\emph{3) Hyperparameters of post-processing}: There are several tunable parameters influence the effectiveness of pseudo-mask refinement.~Table (a) shows results for SAM-enhanced CAM.~Points per side controls how many points SAM samples along each side of the image.
These points act as prompts for SAM,with more points results in finer-grained and more detailed masks.
Table (b) presents the impact of using different strategies for SAM fusion.~The strategies include SAM-AND Fusion, SAM-OR Fusion, and SAM Refinement (Direct Copy).~Table (c) reports results for CRF-based refinement.~Here, the scaling factor plays a critical role in preserving weak yet meaningful activations, preventing them from being overly suppressed during the refinement process.

\begin{table}[htbp]
\centering
\caption{Impact of hyper-parameters and different strategies.~Gray rows indicate best settings.}
\label{tab:hyperparam_post_processing}
\begin{subtable}[t]{0.31\textwidth}
\centering
\begin{tabular}{ccc}
\hline
\textbf{Points} & \textbf{Seed} & \textbf{SAM}  \\ 
\hline
16 & 47.37 &  50.50 \\
\rowcolor{gray!20}
32 & 47.37 &  51.03 \\
\hline
\end{tabular}
\caption{Points per side}
\label{subtab:block}
\end{subtable}
\hspace{0.01\textwidth}
\begin{subtable}[t]{0.31\textwidth}
\centering
\begin{tabular}{ccc}
\hline
\textbf{Strategy} & \textbf{Seed} & \textbf{mIOU} \\
\hline
AND& 47.37 & 41.03 \\
OR& 47.37&  35.97\\
\rowcolor{gray!20}
COPY& 47.37&  51.03\\
\hline
\end{tabular}
\caption{Different strategies for utilizing SAM-generated masks}
\label{subtab:size}
\end{subtable}
\hspace{0.01\textwidth}
\begin{subtable}[t]{0.31\textwidth}
\centering
\begin{tabular}{ccc}
\hline
\textbf{Scaling} & \textbf{Seed} & \textbf{dCRF} \\
\hline
2   &  47.37&  33.47\\
12  & 47.37 &  52.51\\
\rowcolor{gray!20}
16  & 47.37 &  52.52\\
24  & 47.37 & 50.91\\
32  & 47.37 & 48.85\\
\hline
\end{tabular}
\caption{Experiments with CRF parameter}
\label{subtab:momentum}
\end{subtable}
\end{table}

\chapter{Discussion}
\label{chapter:Discussion}
In this section, we start with answering the research questions and insights  learned from the experiments.~After that, the interpretable explanation of the experiment results are presented.

\paragraph{Answer research questions}
Our results in Chapter \ref{chapter:result} shows the effectiveness of our approach.~Based on our experiments, the research questions can be answered as follows:

\noindent\textbf{RQ1}:~Why does the classifier achieve very high accuracy but the activated region of CAM is not accurate or even fail to localize the foreground?

One critical reason is that, ~due to insufficient supervision,~the classification model tends to learn a biased understanding of the foreground.
Under such weak supervision,~it could be very difficult to suppress those diverse background.~Furthermore, the co-occurrence issue is difficult to resolve without additional supervision or explicit causal intervention.~In addition,~the classification activation map typically can only focus on the most discriminative part, ignoring other relevant but less salient areas.

\noindent\textbf{RQ2}:~Is it possible to address co-occurrence issue without external supervision or additional knowledge?

Based on the key observation that complementary knowledge can be extracted from ResNet and ViT,~by adopting the knowledge transfer method with teacher-student paradigm and consistency regularization,~we achieve better feature extraction while mitigating the knowledge bias  without external supervision and additional knowledge. As shown in Chapter \ref{chapter:result}, our proposed method can significantly suppress spurious activations.

\noindent\textbf{RQ3}:~Can we collaboratively  aggregate heterogeneous features from CNN based and Vision Transformer based models to address co-occurrence issue?

By imposing feature consistency across models with distinct architectures via feature alignment techniques,~we achieve effectively knowledge transfer from ResNet to ViT and feature fusion of complementary but heterogeneous features without compromising the strengths of ViT.
% We observe that the integration of feature consistency between heterogeneous backbones ViT and ResNet contributes significantly to suppressing spurious activations.

\paragraph{Training}
Through a series of controlled experiments, we have evaluated the effectiveness of various components in our proposed framework.~Here we will discuss the experimental results and analyze the contribution of each module.

\emph{Comparison with baseline models}:~Due to co-occurrence issue and the inherent weakness of self attention mechanism, ~baseline ViT model's mIOU performance is notably poor, while baseline ResNet model's mIOU is much better. ~Our proposed method is able to address the co-occurrence bias to a large extend,~suggesting that complementary knowledge is able to help the classifier to mitigate spurious correlation and the correctness of our proposed method.~In additional,~We trained the classifier using weak image-level annotations in addition to limited pixel-level annotations.~However,~the model performs poorly and cannot locate the foreground,~suggesting that directly teaching classifier to learn pixel-level knowledge is tricky.~All models are trained for only a few epochs.
Extending training further leads to shortcut learning or overfitting, preventing meaningful performance improvement.

\emph{Comparison with Fully supervised counterparts}:~Our proposed method shows 
competitive results compared to segmentation models that trained with pixel-level annotations,~suggesting that our model has the capacity to be applied in real scenarios.  

\emph{Comparing to Related Works}:~In terms of existing method,~our proposed method outperforms significantly, as our method is able to mitigate the co-occurring issue that occurs quite frequently in our custom dataset. 

\emph{Comparison of teacher-student and co-training paradigm  }:~Table~\ref{tab:paradigm_comparison} shows teacher-student paradigm outperforms co-training paradigm.~The main difference between them is that teacher-student paradigm have a stronger teacher model.
In teacher-student paradigm, the teacher model can provide stable supervision to guide the student model,~whereas co-training relies on mutual learning,~which can result in unstable mutual updates that can amplify noise.

\emph{Why using cosine similarity}:~Table~\ref{tab:paradigm_comparison} shows that using cosine similarity as the metric to compute consistency yields the best performance.~This is because cosine-based loss encourages semantic alignment by measuring directional similarity rather than feature strength. ~In contrast,~other distance metrics like MSE performs poorly in our task, as they measure the magnitude of feature, which can pose unnecessary constraints to the model.

\emph{Feature projection}:~Although feature projection is not indispensable, incorporating this will significantly facilitates the knowledge transfer process by aligning feature distributions and reducing the semantic gap between models.

\emph{Impact of the weighing factor of knowledge transfer loss function}:~Although a small change does not make any significant effect on the result,~we observe that the overall performance is  sensitive to the value of $\lambda$. Therefore,~a well-chosen weighting value is crucial: a low $\lambda$ means that the transfer loss is underutilized, while an overly large $\lambda$ may suppressing other loss components such as classification.

\emph{Dark channel prior}:~In terms of dark channel prior, interestingly, we observe that the background suppression loss  decreases alongside the classification loss, even when training is driven solely by the classification objective. This suggests that the classifier implicitly learns to suppress background noise as it improves its discriminative capability.~However, since  our teacher student framework is already effective in suppressing background noise and achieves great improvement,~and given that our primary goal is to address co-occurrence issue without prior knowledge, thus we do not apply dark channel prior as additional supervision into our pipeline. ~Besides, it is necessary to validate the interaction of different loss function terms before the prior knowledge can be integrated to boost the model's performance.

\paragraph{Post-processing}
Most of the post-processing techniques boosts the performance of the model.~Nevertheless, affinity based method fails to improve the quality of pseudo labels.~Table~\ref{tab:Post-processing} also indicates that combining multiple post-processing techniques does not yield further gains in our setting.~The CAMs  refined through a combination of these methods  do not becoming more matchable with ground truth masks.~This reveals the limitation of post-processing techniques.

\emph{SAM-enhanced}:~Given the precise alignment of SAM masks to object boundaries, we find enhancements as expected in mitigating partial and false activations in the existing pseudo-labels.

\emph{CLIP-aided}:~Despite experimenting with various text prompts, our method failed to effectively distinguish smoke from non-smoke images.~This is mainly due to the difficulty in formulating appropriate prompts that can clearly discriminate between smoke and non-smoke regions.~This also suggests that CLIP’s pre-trained visual-textual representations  lack the knowledge required to capture  visual differences in our domain-specific task.
% due to the difficulty in formulating appropriate prompts that can clearly discriminate between smoke and non-smoke regions.

\emph{Cam fusion}:~For cam fusion,~significant improvements are observed in Table~\ref{tab:cam_impact} when the initial seed is of low quality.
By fusing features across layers, we improve the robustness of CAMs against localization and classification errors, leading to more accurate and complete pseudo-labels.~However, when the initial seed is already strong, incorporating shallow features may introduce noise, which can negatively affect the refinement process.~This in another way also showcasing that with our framework, partial coverage problem can also be alleviated to some extend, reducing reliance on some post-processing techniques.

\emph{Affinity-based}:~This kind of method fails to refine the pseudo mask.~The reason is that affinity propagation tends to smooth labels across similar regions in a class-agnostic manner,~lacking of high-level semantics.~As a result,~it may wrongly activate background regions especially for low contrast foreground like smoke.  

\chapter{Conclusion and Future Work}
\label{chapter:conclusion}

\section{Conclusion}
Because the absence of localization information,~weakly supervised semantic segmentation from image-level supervision suffers from several main challenges, such as incomplete foreground coverage and co-occurrence bias.
In this work,~we investigate how the complementary strengths of Vision Transformers and ResNet architectures can be jointly utilized to reduce knowledge bias in order to mitigate spurious correlations.~Unlike previous methods that rely heavily on external supervision or human prior, our proposed framework leverages the rich semantic and contextual representations captured by ViT and ResNet to achieve internal supervision.~By enforcing feature-level consistency between heterogeneous models, our proposed method effectively  suppress spurious activations and improves boundary localization, particularly for ambiguous object like smoke.~Moreover,~most previous  WSSS approaches have been limited to built upon a single type of network architecture for feature extraction, our method introduces a new perspective by integrating heterogeneous features through consistency regularization to mitigating spurious correlations.
% This also highlights the potential of cross-architecture consistency as a promising direction for enhancing pseudo-label quality without requiring additional supervision.

In conclusion, we propose a novel WSSS framework based on teacher-student paradigm to tackle a number of core challenges in weakly supervised semantic segmentation.
Our framework not only reduces dependency on pixel-level annotations, but also bridges the gap with fully supervised semantic segmentation method.
By enforcing feature-level consistency between architectures, we reduce co-occurring bias and achieve more accurate boundary precision for ambiguous objects like smoke.~To mitigate partial coverage  and low activation issues,~we also explore a bunch of advance post-processing techniques to further enhance the quality of pseudo labels.

\section{Limitations and Future Work}
However, due to time constraints, some interesting and promising parts remain underexplored.~In future work, we will further investigate the effectiveness of
our framework to other datasets.~Furthermore, we aim to investigate more sophisticated knowledge transfer techniques, improving feature alignment techniques to better bridge the gap between heterogeneous architectures.~In addition, we plan to integrate advanced mask refinement strategies for post-processing using SAM.~Another key direction is to  directly perform weakly supervised video segmentation  on our videos instead of converting them to static pictures.~Finally, we intend to evaluate our proposed methods in more complex settings--from binary class to multiclass and from multi stage to end to end approach.

% \newpage
% \setcounter{secnumdepth}{-1}
% \input{content/acknowledgement}

%% Prevent urls running into margins in bibliography
\setcounter{biburlnumpenalty}{7000}
\setcounter{biburllcpenalty}{7000}
\setcounter{biburlucpenalty}{7000}

%%  Add the bibliography
\printbibliography[heading=bibintoc,title=References]

% Letters for chapters
\appendix

\setcounter{secnumdepth}{2}

\end{document}